\documentclass[conference]{IEEEtran}

\usepackage[percent]{overpic}
\usepackage{booktabs}
\usepackage{subfiles}
\usepackage{cite}
\usepackage{amsmath,amssymb,amsfonts, amsthm}
\usepackage{algorithmic}
\usepackage{graphicx}   
\usepackage{textcomp}
\usepackage{xcolor}
\usepackage{comment}
\usepackage{tabularx}
\usepackage{caption,subcaption,graphicx}
\usepackage{multirow}
\usepackage{subcaption}
\usepackage[font=small,labelfont=bf]{caption}
\usepackage{float}
\usepackage{adjustbox} 
\usepackage{placeins}
\definecolor{darkgreen}{rgb}{0.0, 0.5, 0.0}
\def\BibTeX{{\rm B\kern-.05em{\sc i\kern-.025em b}\kern-.08em
    T\kern-.1667em\lower.7ex\hbox{E}\kern-.125emX}}
\begin{document}
\renewcommand{\thesubfigure}{\Alph{subfigure}}
\captionsetup[subfloat]{labelfont = {large, bf}, format = plain, labelformat = simple, labelsep = period, singlelinecheck = false}
%\title{On Neuronal Ensemble to Ensemble Communication in Artificial Neural Networks}
\title{Neuro-inspired Ensemble-to-Ensemble Communication Primitives for Sparse and Efficient ANNs}

\author{
\IEEEauthorblockN{
Orestis Konstantaropoulos\IEEEauthorrefmark{1}, 
Stelios Manolis Smyrnakis\IEEEauthorrefmark{2}, 
Maria Papadopouli\IEEEauthorrefmark{1}\IEEEauthorrefmark{3}\IEEEauthorrefmark{4}}
\\
\IEEEauthorblockA{\IEEEauthorrefmark{1}Archimedes, Athena Research Center, Athens, Greece\\
\IEEEauthorrefmark{2}Department of Neurology, Brigham and Women’s Hospital, Harvard Medical School, Boston, USA\\\IEEEauthorrefmark{3}Department of Computer Science, University of Crete, Greece\\\IEEEauthorrefmark{4}Institute of Computer Science, Foundation for Research and Technology-Hellas, Greece}}

%\author{\textbf{Anonymous Authors}}

\maketitle
%%%%%%%%%%%%%%%%%%%%%%%%%%%%

\begin{abstract}
The structure of biological neural circuits—modular, hierarchical, and sparsely interconnected—reflects an efficient trade-off between wiring cost, functional specialization, and robustness. These principles offer valuable insights for artificial neural network (ANN) design, especially as networks grow in depth and scale. Sparsity, in particular, has been widely explored for reducing memory and computation, improving speed, and enhancing generalization. Motivated by systems neuroscience findings, we explore how patterns of functional connectivity in the mouse visual cortex—specifically, ensemble-to-ensemble communication—can inform ANN design. We introduce G2GNet, a novel architecture that imposes sparse, modular connectivity across feedforward layers. Despite having significantly fewer parameters than fully connected models, G2GNet achieves superior accuracy on standard vision benchmarks. To our knowledge, this is the first architecture to incorporate biologically observed functional connectivity patterns as a structural bias in ANN design. We complement this static bias with a dynamic sparse training (DST) mechanism that prunes and regrows edges during training. We also propose a Hebbian-inspired rewiring rule based on activation correlations, drawing on principles of biological plasticity. G2GNet achieves up to 75\% sparsity while improving accuracy by up to 4.3\% on benchmarks, including Fashion-MNIST, CIFAR-10, and CIFAR-100, outperforming dense baselines with far fewer computations.
\end{abstract}

\begin{IEEEkeywords}
Neuronal Ensembles, Sparsity in Deep Learning, Functional Connectivity, Computer Vision
\end{IEEEkeywords}

\section{Introduction}
Neuroscience has been a significant source of inspiration in the design of deep learning, for addressing challenges, such as energy cost, ability for generalization and continual learning. Concepts such as spiking neural networks and neuromorphic hardware take cues from the event-driven, sparse firing patterns observed in the brain to enable low-power, efficient computation \cite{snn1, snn2, snn3}. 
Biologically plausible learning rules, motivated by Hebbian learning and spike-timing-dependent plasticity (STDP), offer promising alternatives to backpropagation for unsupervised or local learning \cite{dfa, ffa, ff, pepita, softheb}.
%Although these approaches have not yet matched state-of-the-art results on large-scale benchmarks, they offer promising directions for scalable, unsupervised learning. Additional work incorporates 
Biologically-inspired components, such as precalibrated Gabor filter banks, lateral inhibition, and winner-take-all dynamics, have been incorporated in ANNs to improve robustness against adversarial perturbations and enhance interpretability \cite{vonenet, theod}.
The structure of biological neural circuits, e.g., its modular, hieararchical, sparse inter-connectivity--thought to reflect an efficient trade-off between wiring cost, functional specialization, and robustness--offers further insight for the design of ANNs. 
The influence of such connectivity principles on ANNs has been increasingly explored \cite{efficient_sparse, brains, exploring, wiring, italian, snns_brain, generative,sporns2016networks,SPORNS200655,barabasi, doro}. Sparsity, in particular, is both a biological necessity and a desirable property in artificial systems. As artificial neural networks (ANNs) continue to expand in depth and size \cite{scale}, sparsity has been widely studied as a means to reduce memory footprint, computational cost, and often training and inference time \cite{sparse2, spars, structured_spars, lora}. Moreover, sparsity may not only improve the efficiency of artificial networks but also their ability to generalize by promoting more compact internal representations and improving robustness to noise.

While the properties of individual biological neurons are well characterized, our understanding of how cortical neuron networks coordinate to process information remains limited. \textit{Neuronal ensembles} that exhibit \textit{synchronous firing} are thought to transmit shared information to downstream targets more efficiently \cite{10.1162/NECO_a_00476} and have been highlighted in several seminal studies examining both spontaneous and stimulus-evoked activity patterns \cite{de1949cerebral, %Meister1991, 
Abeles1993, Grinvald2003, hebb1949first, Ringach2009, CarrilloReid2016,panzeri2022structures,averbeck2006neural, yuste:2015, Cossart:2003Attractor,mil,russo2017cell,Yuste2024}.
Putative communication primitives that mediate neuronal activity and information transmission, across cortical layers, further highlight the functional role of sparsity in information processing and communication\cite{olshausen}. 
For example, a recent study \cite{pap} found that biological neurons, along with their functionally-connected partners (i.e., other neurons with which they exhibit \textit{statistically-significant pairwise synchronous firing}), form elementary \textit{multi-neuronal ensembles (modules)} that act as core information-processing units both within and across granular and supragranular layers in mouse primary visual cortex. 
Layer 4 is considered as the primary input layer of the cortex, while supragranular layers Layer 2 and Layer 3, receiving primarily input from L4, integrate and relay information to other cortical areas.
These neural ensembles exhibit strong
cooperativity, with \textit{aggregate cofiring} activity—rather than \textit{individual neuron identity}—driving the L2/3 neuron
responses in a ReLU-like manner. In particular, across cortical laminae, the firing probability of Layer 2/3 pyramidal neurons exhibits a \textit{ReLU-like activation} profile as a function of the number of its L4 functionally-connected neurons. Notably, the firing probability of a Layer 2/3 neuron increases \textit{sharply} when at least approximately 13\% of its Layer 4 partners co-fire, a relatively rare event, reflecting a\textit{ nonlinear integration mechanism} that promotes both \textit{reliable supra-threshold transmission} and \textit{sparse activity}. Moreover, ensemble-to-ensemble information transfer from Layer 4 to Layer 2/3 demonstrates enhanced specificity, sensitivity, accuracy, and precision. We speculate that such ensembles (modules),
being \textit{flexible} and \textit{multiplexing}, may support task-specific processing, adaptation, and learning through
dynamic reconfiguration of their interactions.
\begin{comment}
Here we aim to use examine the role of these pathways, ensemble to ensemble communication, by simulating the paradigm in ANNs and SNNs, on their performance. Moreover, we will apply our methodology that identifies the functional connectivity in these ANNs, to assess they exhibit similar principles of communication as the ones observed in biological networks.
To our knowledge, it is the first time that functional connectivity patterns identified in the 
mice visual cortex are integrated in ANNs to assess their impact on the performance of the artificial architecture.
\end{comment}

Inspired by this work \cite{pap}, we aim to examine the role of these "pathways", i.e., ensemble-to-ensemble communication primitives, by imposing a similar structure on the feedforward layers of ANNs. The proposed architecture, \textbf{G2GNet}, introduces sparse, modular connectivity across layers. This approach allows us to analyze the role of functional connectivity structures observed in mice visual cortex while simultaneously proposing an efficient deep learning architecture. Despite having significantly fewer parameters and computational load compared to a fully-connected one, G2GNet achieves superior performance on computer vision benchmarks. To our knowledge, this is the \textit{first} study that integrates \textit{functional connectivity} patterns and organization principles identified in the visual cortex as a \textit{structural bias into ANNs} to assess their effect on the performance of artificial architecture.
%
%Inspired by group-to-group communication observed in cortical ensembles, 
Specifically, each layer in G2GNet is partitioned into groups that \textit{preferentially connect} to corresponding groups in adjacent layers. 
We complement this static structural bias with a dynamic sparse training (DST) mechanism that prunes \textit{and} regrows edges \textit{during training}. Furthermore, we introduce a Hebbian-based criterion for edge updates, guided by activation correlations among neurons—inspired by the plasticity observed in biological neural networks.

In summary, this work makes the following key contributions: \textbf{1)} Develops an ANN architecture that incorporates neuronal grouping strategies inspired by the functional connectivity patterns of biological neural networks as a \textit{structural bias}; \textbf{ 2)} Integrates a Hebbian-inspired criterion for\textit{ dynamic addition} and \textit{removal of connections} in the proposed architecture; \textbf{ 3)} Analyzes the impact of different layer-wise grouping strategies on network accuracy; \textbf{4)} Demonstrates that the proposed structural bias achieves up to \(75\%\) sparsity in feedforward layers while improving accuracy by up to \(4.3\%\) on computer vision benchmarks, such as Fashion-MNIST, CIFAR-10 and CIFAR-100, compared to fully-connected architectures.
\begin{figure*}[t]
    \centering
    \includegraphics[width=1\linewidth]{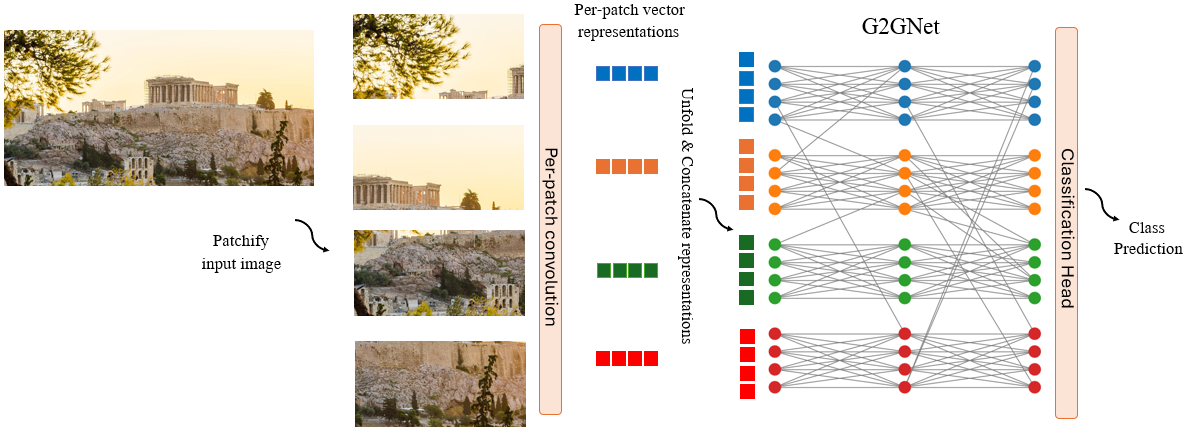}
    \caption{\textbf{Overview of the proposed G2GNet architecture.} The input image is first divided into patches (four in this example), and a convolutional layer is applied \textit{independently} to each patch to produce vector representations. These vectors are concatenated and passed through a feed-forward network with a predefined \textit{sparse structure}, inspired by\textit{ group-to-group} (neural ensemble to ensemble) communication patterns observed in biological neural circuits. We refer to this architecture as \textbf{G2GNet}. A final linear classifier predicts the class of the input image. Here, for clarity, we demonstrate a simplified instance of G2GNet where each layer has 16 nodes divided in 4 groups.}

    \label{fig:overview}
\end{figure*}

\section{Background: Sparsity in Deep Learning}\label{sec:back}

One of the most common approaches to \textbf{compress} neural networks is through weight pruning after training~\cite{lecun_opt, group_lasso, importance, state, hoefler}. A key insight from this line of work is provided by the Lottery Ticket Hypothesis~\cite{lottery, lot2}, which suggests the existence of \textbf{sparse subnetworks}, “winning tickets”, that when \textit{trained} in isolation, can match the performance of their dense counterparts. That is, a “winning ticket” is a subnetwork that performs
as well as the full network.
To bypass the need for training large, dense models to uncover such subnetworks, alternative methods have been proposed that initialize and train sparse networks directly~\cite{snip, pick_w}. 
These approaches fall under the umbrella of \textit{static sparse training}, as the set of active edges remains \textit{fixed} throughout the training process.

In contrast, \textbf{dynamic sparse training} (DST) explores methods for updating the sparse connectivity of a network during training~\cite{Mocanu, mostafa, bellec, riggl, topk}. DST \textit{dynamically} prunes and regrows connections based on specific criteria, allowing the network structure to evolve, while maintaining a constant number of edges. For example, the Sparse Evolutionary Training (SET) method~\cite{Mocanu} removes weights with the smallest magnitudes, while other approaches~\cite{dettmers2020sparse, riggl} incorporate gradient information to guide the update process, often at the expense of increased memory and computational cost. Inspired by neuroplasticity, DST offers a framework for topological flexibility, enabling the model to adapt its structure. This process mirrors the synaptic regeneration and rewiring in the brain as well as the dynamic “reorganization" of the functional-connectivity observed in visual cortex \cite{pap}. 

Apart from the distinction between \textit{dynamic} and \textit{static sparsity}, sparsity can also be categorized as either \textbf{structured} or \textbf{unstructured}.
In unstructured sparsity, individual weights (connections) are pruned \textit{independently }based on a given criterion, commonly magnitude-based pruning. While this typically leads to irregular sparsity patterns (i.e., individual elements in the weight matrix are zeroed out), the overall structure and dimensions of the layers remain unchanged. This form of sparsity offers high theoretical compression but is difficult to exploit for actual speedups on general-purpose hardware (e.g., CPUs or GPUs) without specialized sparse libraries or hardware support~\cite{hoefler}.

In contrast, structured sparsity applies \textit{pruning at higher levels of abstraction}, such as channels (feature maps), filters (entire convolution kernels), neurons (rows in fully connected layers), or even entire layers (e.g., residual blocks in ResNets)\cite{group_lasso, explo}. This results in more \textit{regular and hardware-friendly sparsity patterns} that can be efficiently accelerated using standard hardware. However, achieving high accuracy under such constraints is often more difficult\cite{hw_prune}.

Among structured sparsity techniques, \textbf{$N$:$M$} sparsity has received particular attention due to its compatibility with modern accelerators~\cite{nm}. In this scheme, every block of $M$ elements in the weight matrix contains exactly $N$ \textit{non-zero values}, a property also shared by the connectivity patterns in G2GNet. Recent work has also explored how to dynamically train $N$:$M$ sparse networks using DST~\cite{strdst}.

%While \textit{unstructured sparsity} offers strong compression and accuracy trade-offs, its irregular patterns can limit the efficiency gains achievable on real hardware. To address this, \textbf{structured sparsity} methods impose constraints on the sparsity pattern, such as\textit{ block} or \textit{channel-level} pruning, i.e., removal of entire channels (feature maps) from convolutional layers~\cite{group_lasso, explo}. Although this can sometimes degrade accuracy compared to \textit{unstructured} methods, it enables more predictable and hardware-friendly acceleration. Among these, $N$:$M$ sparsity—where each block of $M$ weights contains exactly $N$ non-zero values—has gained attention for its compatibility with modern accelerators, and recent work has investigated how to train such structured patterns dynamically using DST~\cite{strdst}.

Our work builds on the growing body of research of brain-inspired deep learning and sparsity in ANNs by incorporating functional connectivity patterns from biological networks as a \textit{structural bias in ANN architectures}. Our sparse architecture builds on\textit{ neuronal ensembles, i.e., groups} (used interchangeably in the paper) \textit{within each layer} and their \textit{inter-layer communication}. This introduces the notion of \textbf{pathways}: active groups that transmit information between layers, bringing \textit{modularity} and \textit{structured sparsity} into the design. In addition, we explore the integration of \textit{dynamic} sparse training techniques within this biologically informed framework. We introduce a \textit{Hebbian-like criterion} for pruning and growing connections inspired by biological synaptic plasticity.

% In summary, this work makes the following key contributions: \textbf{1)} The development of an ANN architecture that incorporates functional connectivity patterns from biological networks as a structural bias; \textbf{ 2)} The integration of a hebbian-inspired criterion for the dynamic addition and removal of connections in the proposed architecture. \textbf{ 3)} The analysis of the impact of different layer-wise grouping strategies on network accuracy; \textbf{4)} We showed that the proposed structural bias achieves up to \(75\%\) sparsity in feedforward layers while improving accuracy by up to \(4.3\%\) on computer vision benchmarks compared to fully-connected architectures.

%\section{Experiments, Data Collection, and Pre-processing}\label{sec:exp}
%
%

\section{Methodology}\label{sec:meth}
We will examine the role of the ensemble-to-ensemble information flow observed in biological networks by imposing a structural bias on the feedforward layers of ANNs.
A key challenge is how the functional grouping observed in biological neurons can be translated into \textit{structural grouping in ANNs}. We propose that \textbf{grouping} can be achieved by ensuring that\textit{ neurons within the same group} \textit{share highly similar connectivity patterns}, that is, neurons in a specific group (say \textit{i}) connect to the \textit{same neurons} in the next layer with the \textit{same probability}. Additionally, each group is \textit{strongly coupled with its corresponding group in the next layer}, while maintaining \textit{only sparse connections with other groups} in that layer.
%pairs of groups in \textit{consecutive layers }are \textit{strongly coupled}, exhibiting high inter-connectivity. 
This design results in a highly sparse network with small-world-like properties across its layers. In the following paragraphs, we present the details of our architecture.

\subsection{Sparse layer structure inspired by primary visual cortex (V1)}

Inspired by the modular organization and group-specific connectivity observed in the primary visual cortex (V1), we propose a sparse architecture that enforces structured modularity in feedforward layers. Conceptually, our \textbf{G2GNet} architecture defines a \textit{probabilistic connectivity mask} between consecutive layers that activates \textit{only} a subset of the possible connections, favoring \textit{within strongly coupled groups communication.}

Let the number of neurons in layer \(i\) be \(N_i\), and in layer \(i+1\) be \(N_{i+1}\). We formally define the construction of a layer's connectivity as follows:
\begin{enumerate}
    \item Partition the \(N_i\) neurons of layer \(i\) into \(G_i\) groups, each containing \(N_i / G_i\) neurons. 
    % Similarly, 
    \item Partition the neurons of layer \(i+1\) into \(G_{i+1}\) groups.
    \item For each neuron in group \(k\) of layer \(i\), connect it to neurons in group \(k\) of layer \(i+1\) with probability \(\textbf{p}\), and to neurons in\textit{ all other groups} of layer \(i+1\) with a \textit{much lower probability} \(\textbf{p}' \ll \textbf{p}\).
\end{enumerate}

\begin{figure}[t]
    \centering
    \includegraphics[width=0.7\linewidth]{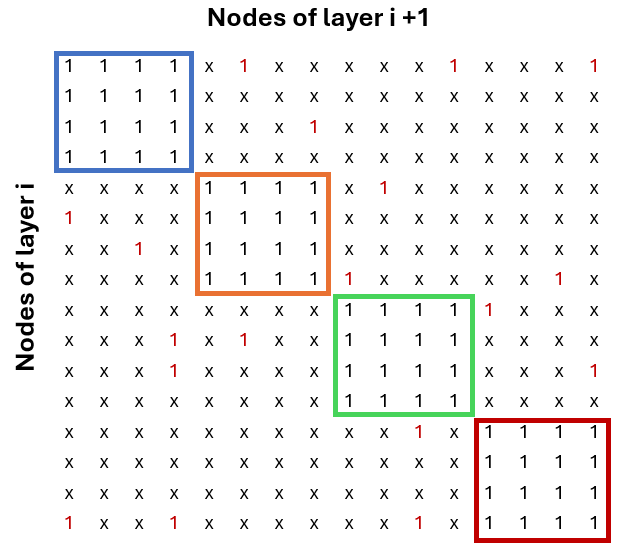}
    \caption{\textbf{Adjacency matrix between two consecutive layers in G2GNet.} Assume layers \(i\) and \(i+1\) each contain 16 neurons, partitioned into 4 groups. Neurons in group \(k\) of layer \(i\) are connected to neurons in group \(k\) of layer \(i+1\) with high intra-pathway probability (\(p = 1\)), and to neurons in other groups with low inter-pathway probability (\(p' = 0.05\)). A value of 1 in the adjacency matrix indicates an \textit{active connection}; otherwise, the entry is marked as \(x\). The resulting matrix exhibits an approximately block-diagonal structure, reflecting the \textit{group-to-group} connectivity pattern.}

    \label{fig:adj}
\end{figure}

This construction results in a connectivity matrix between layers \(i\) and \(i+1\) that is approximately \textit{block diagonal}: the \textit{diagonal} blocks, corresponding to same-index groups, are \textit{dense} with probability \(\textbf{p}\), while the \textit{off-diagonal blocks} are much \textit{sparser} with probability \(p'\). Figure~\ref{fig:overview} schematically depicts the G2GNet architecture and Figure~\ref{fig:adj} visualizes the resulting connectivity mask given a specific instance. Both Figures depict a simplified case where \(G_i = 4, N_i = 16\) for each layer \(i\). The hyperparameters used in our experiments are noted in Section \ref{sec:exp}.

As information flows across layers, it primarily travels along pathways formed by densely-connected group pairs that are connected with a high probability \(p\). In parallel, information also propagates through significantly sparser connections between the groups, i.e., connected with the lower probability \(p'<<p\), allowing for secondary information leakage across pathways. For clarity, throughout the rest of the paper, we refer to \(p\) as \textit{intra-pathway probability} and \(p'\) as \textit{inter-pathway probability}.

The proposed pattern can be highly sparse. Instead of maintaining all \(N_i \times N_{i+1}\) parameters, the number of active connections becomes
\[
N_i \times N_{i+1} \times \left[ \frac{p}{G_{i+1}} + \frac{p' (G_{i+1} - 1)}{G_{i+1}} \right].
\]
For example, with \(p = 1\), \(p' = 0.05\), and \(G_i = G_{i+1} = 8\), the architecture achieves 85\% sparsity, utilizing only 15\% of the potential parameters.

One might ask why not simply retain a random 15\% of connections rather than enforcing a modular structure. However, inspired by biological observations of functional neuronal groups, our structured connectivity consistently yields higher accuracy on vision datasets, see \ref{sec:exp}. In addition, the explicit group structure enables \textit{structured sparsity}, which is more compatible with acceleration on modern hardware.

\subsection{Grouping strategies}
\label{sec:gr}
The way neurons are grouped within each layer plays a crucial role in \textit{shaping how information flows} through the network. Conceptually, grouping determines which neurons are densely connected and which interact only sparsely, thereby influencing both spatial locality and feature integration.  

\begin{figure*}[t]
    \centering

    % Subfigure a
    \begin{subfigure}[b]{0.3\linewidth}
        \centering
        \includegraphics[height=4cm]{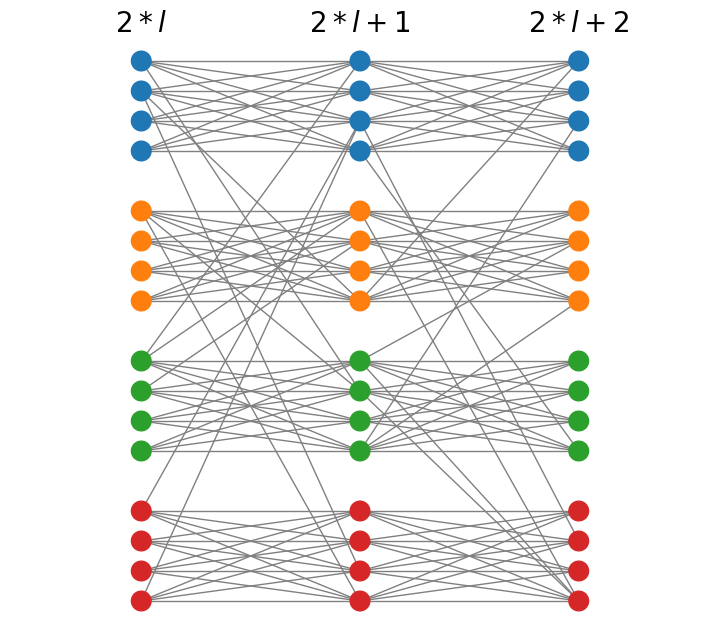}
        \caption{\textbf{Index-based}}
        \label{fig:index}
    \end{subfigure}
    \hfill
    % Subfigure b
    \begin{subfigure}[b]{0.3\linewidth}
        \centering
        \includegraphics[height=4cm]{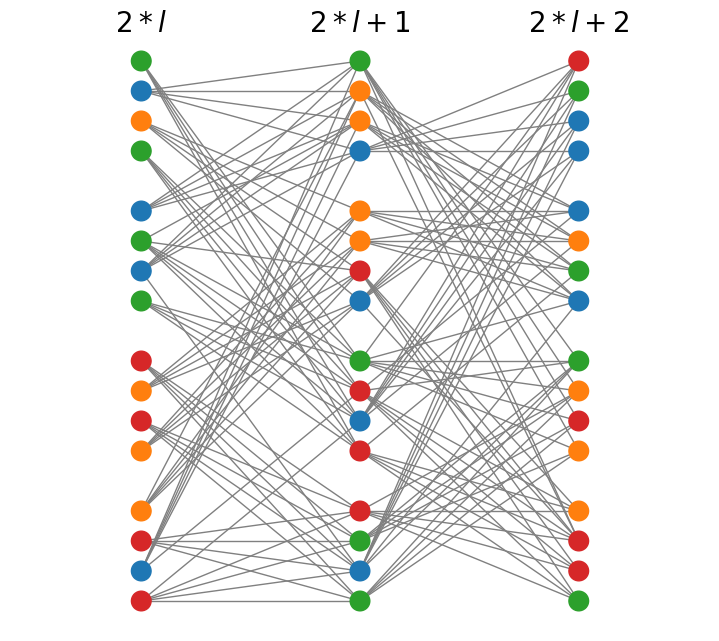}
        \caption{\textbf{Random}}
        \label{fig:random}
    \end{subfigure}
    \hfill
    % Subfigure c
    \begin{subfigure}[b]{0.3\linewidth}
        \centering
        \includegraphics[height=4cm]{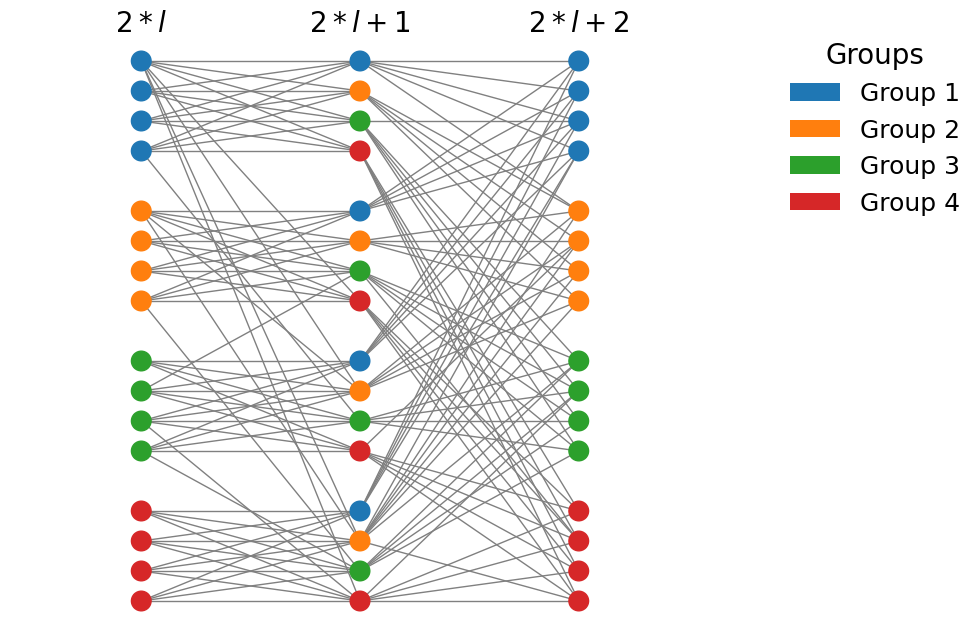}
        \caption{\textbf{Mixer}}
        \label{fig:mixer}
    \end{subfigure}

    \vspace{1em}
    \caption{
    \textbf{Comparison of neuron grouping strategies across three consecutive layers of the G2GNet.}  
    Each subfigure illustrates three consecutive layers, with neurons color-coded by group membership. \textit{Note that groups are defined such that all neurons within a group share the same connection probabilities to neurons in the next layer.} We demonstrate a simplified instance of G2GNet where \(G_i = 4, N_i = 16\) for each layer \(i\).  
    \textbf{(a) Index-based grouping:} neurons are divided into contiguous blocks by index, preserving spatial locality from the input image across layers.  
    \textbf{(b) Random grouping:} neurons are assigned to groups without regard to position, disrupting spatial structure and coherence.  
    \textbf{(c) Mixer strategy (proposed):} groups alternate between index-based (even-numbered layers) and interleaved assignments (odd-numbered layers, where group \(k\) contains all neurons \(l\) such that \(k = l \mod G_i\)). This approach maintains spatial locality in some layers while promoting integration across spatial regions in others—balancing modularity and feature mixing.
}

    \label{fig:grouping_strategies}
\end{figure*}

Several grouping strategies can be considered: neurons can be grouped by their index (``\textbf{Index-based}", Fig. \ref{fig:index}), preserving the ordering induced by the input. Alternatively, neurons can be assigned to groups \textit{randomly}, disregarding any structure (``\textbf{Random}", Fig. \ref{fig:random}). In this work, we propose a \textit{hybrid} method inspired by the MLPMixer architecture \cite{mlpixer}, which we call the \textit{Mixer} strategy (``\textbf{Mixer}" Fig. \ref{fig:mixer}). The Mixer \textit{alternates} between two grouping schemes in successive layers:  
\begin{itemize}
    \item In even-numbered layers, neurons are partitioned by index into consecutive blocks, as described previously.
    \item In odd-numbered layers, neurons are assigned to groups in an interleaved fashion: each neuron with index \(l\) is assigned to group \(k\) such that \(k = l\mod G_i\).  
\end{itemize}

Figure~\ref{fig:grouping_strategies} illustrates this alternating grouping mechanism, again in the simplified case where \(G_i = 4, N_i = 16\) for each layer \(i\).
\captionsetup[subfigure]{justification=centering}

The choice of grouping affects \textit{how spatial information from the input image is processed}. When neurons are grouped by index, \textit{spatial proximity of the input pixels is preserved across layers}: neurons within the same group predominantly process information originating from the same spatial region of the input, while inter-pathway (sparser) connections enable \textit{integration from more distant regions}. This index-based grouping is loosely inspired by the columnar structure in the higher mammals of neurons with similar properties, like orientation columns. In contrast, random grouping destroys this spatial locality, as neurons from different regions of the input are arbitrarily combined.  

The proposed Mixer strategy strikes a balance: even-numbered layers preserve spatial locality by grouping neurons that process nearby input regions, while odd-numbered layers promote integration of information from different regions by mixing neurons from distinct spatial areas into the same group. Using MLPMixer terminology, in even layers groups correspond to \textit{patches}, while in odd layers groups correspond to \textit{channels}. Unlike the original MLPMixer, however, we allow communication both within and between patches and channels in every layer through the low-probability inter-pathway \(p'\) connections.

\subsection{Dynamic Sparse Training}

Having defined the G2GNet architecture and imposed its biologically inspired structural bias, we now introduce a mechanism to \textit{dynamically update the network’s sparse connectivity} during \textit{training}, following the principles of Dynamic Sparse Training (DST).
A common approach in DST is to initialize the network using the Erdős–Rényi (ER) method, which generates a random binary mask for each layer to enforce a predefined sparsity level. \textit{During training}, the sparse connectivity is \textit{periodically updated} every \(\Delta t\) training iterations. A fraction of the currently \textit{active} edges is removed according to a pruning criterion \(C_P\), rendering them inactive and excluding them from the forward and backward passes. An \textit{equal number} of connections is then reintroduced (added) from the pool of inactive edges based on a growth criterion \(C_G\), maintaining a constant overall sparsity level. Below we describe the pruning and growth (addition) criteria.
\subsection{Rewiring Criteria during Training}
We explore several combinations of pruning and growth criteria to apply to G2GNet. For pruning, we consider:  
\textbf{(i)} a backpropagation (BP) magnitude-based criterion that removes edges with the smallest (BP) weight, as in \cite{Mocanu, fantastic},  
\textbf{(ii)} random selection, and  
\textbf{(iii) }a novel Hebbian-inspired criterion \(C_H\), described below.  
For growth, we examine the random addition and the Hebbian-based criterion \(C_H\). All pruning and regrowth steps are applied every \(\Delta t = 1000\) iterations, dropping and adding \(2.5\%\) of the edges in each layer. A detailed sensitivity analysis, varying these hyperparameters, is presented in Section~\ref{sec:exp}.

%During training, we remove a fraction of the \textit{active} weights based on a pruning criterion  \(C\), rendering them \textit{inactive} and excluding them from the network’s computation. Then, an \textit{equal number} of connections are reintroduced (regrown) from the pool of inactive weights according to a growth criterion—\textit{ensuring that the overall sparsity level remains constant.}

We propose a \textit{Hebbian-inspired update mechanism} tailored to the modular structure of G2GNet. Specifically, we introduce a criterion \(C_H\) that adds or removes connections based on the functional connectivity of intergroup neurons. The idea is to favor connections between neurons whose activations are more strongly correlated, in line with Hebbian plasticity principles.

Let \(x_i, x_j \in \mathbb{R}^{B}\) denote the activation vectors of neurons \(i\) and \(j\) over a batch of size \(B\). We define the pairwise \textbf{functional connectivity of neurons \(i\) and \(j\) }during a period as the cosine similarity between their activation vectors during that period:
\begin{equation}
    C_H(i, j) = \frac{x_i x_j^\top}{\|x_i\|_2 \cdot \|x_j\|_2}
\end{equation}
where \(\|\cdot\|_2\) denotes the \(L_2\) norm. 
When used as a pruning criterion, edges with the lowest functional connectivity scores are removed. Conversely, when used as a growth criterion, \textit{new edges are added} by selecting those with the highest functional connectivity scores from the set of \textit{inactive} connections.

\begin{comment}
    
\subsection{Structured sparsity using a router network}  

We now explore the benefits of our proposed architecture in enabling structured sparsity. Thanks to its modular design, we can dynamically deactivate a subset of the dense blocks during inference without significantly impacting performance, thereby reducing the number of computations. This also reinforces the notion of \textit{pathways}, where only a subset of ensembles remains active depending on the input image. Conceptually, this idea aligns with the \textit{Mixture of Experts} paradigm used in state-of-the-art large language models (LLMs), where the top-\(k\) experts are selectively activated in a data-dependent manner.  

We find empirically that even randomly dropping a fixed number of groups during inference performs surprisingly well. However, for more data-driven and accurate selection, we introduce an external \textit{router network}. This lightweight network takes as input the activations of a layer and predicts which groups should remain active in the next layer. The router itself accounts for roughly 10\% of the computations of the corresponding feedforward layer and enables masking of up to half of the groups, leading to an overall reduction of about 40\% in computational cost.
\end{comment}
\begin{table*}[ht]
\centering
\caption{
Test accuracies (\%) of different connectivity patterns on FashionMNIST (F-MNIST), CIFAR-10 and CIFAR-100. 
We compare fully-connected architectures with the same number of parameters (v1) or same width (v2) as ours, 
as well as ER random sparse networks and our V1-inspired topologies.
e}
\label{tab:primary_tab}
\begin{tabular}{lcccc}
\toprule
\textbf{Connectivity Pattern} & \textbf{F-MNIST Accuracy (\%)} & \textbf{CIFAR-10 Accuracy (\%)} & \textbf{CIFAR-100 Accuracy (\%)} & \textbf{Num. Parameters} \\
\midrule
\textbf{Fully-Connected v1} & 90.2 & 68.4 & 38.7 & 806,400 \\
\textbf{Fully-Connected v2} & 90.9 & 71.5 & 40.8 & \textit{3,150,000} \\
\textbf{ER Random Graph} & 89.2 & 69.8 & 40.7 & 806,100 \\
%\textbf{G2GNet (Index Grouping, Ours)} & 90.4 & 70.9 & 42 & 806,100 \\
\textbf{G2GNet (Proposed)} & \textbf{91.14} & \ \textbf{71.5} & \textbf{42.2} &806,100 \\
\bottomrule
\end{tabular}
\end{table*}
\section{Experiments}
\label{sec:exp}
\subsection{Architectural Details}
To evaluate the effectiveness of the proposed G2GNet architecture, we conduct experiments on three standard image classification benchmarks: Fashion-MNIST and CIFAR-10 each contain 10 classes, while CIFAR-100 includes 100 classes. Fashion-MNIST consists of grayscale images of size \(28 \times 28\), whereas CIFAR-10 and CIFAR-100 contain RGB images of size \(32 \times 32\). These datasets are widely used for evaluating the performance of novel neural network architectures under varying conditions. Although G2GNet is applied in a purely feedforward setting, and thus results on CIFAR-100 are naturally lower, they still provide informative insights into the effectiveness of the approach.

Unless otherwise stated, we use the following default hyperparameters throughout our experiments: \(p = 1\), \(p' = 0.15\), and \(G_i = 8\) for all layers. The G2GNet consists of three hidden layers, each with 1024 neurons with ReLU activations. The input image is divided into 16 patches, and a convolutional layer with 32 output channels is applied to each patch independently. The resulting patch embeddings are concatenated and passed through the G2GNet. Given this configuration, each group in the first hidden layer corresponds to two input patches. A final linear classifier produces the output logits.

The model is trained end-to-end for 20 epochs using backpropagation with cross-entropy loss. Optimization is performed using the Adam optimizer with a learning rate of \(1 \times 10^{-3}\).

\begin{figure*}[ht]
    \centering
    \begin{subfigure}[b]{0.21\textwidth}
        \centering
        \caption{}
        \includegraphics[width=\linewidth]{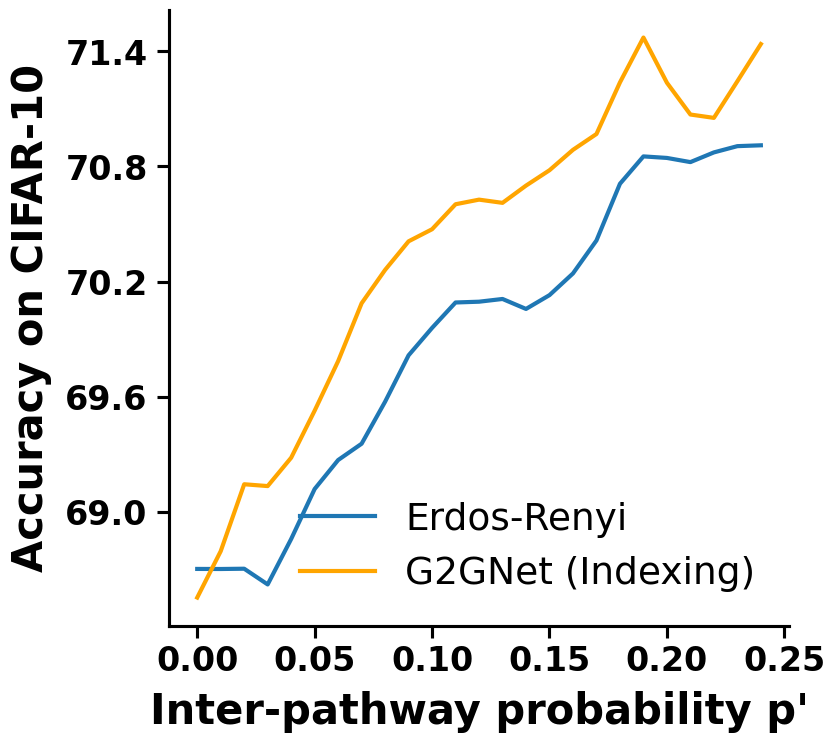}
        %\caption{Connectivity patterns.}
        \label{fig:er_pprime}
    \end{subfigure}
    \hfill
    \begin{subfigure}[b]{0.19\textwidth}
        \centering
        \caption{}
        \includegraphics[width=\linewidth]{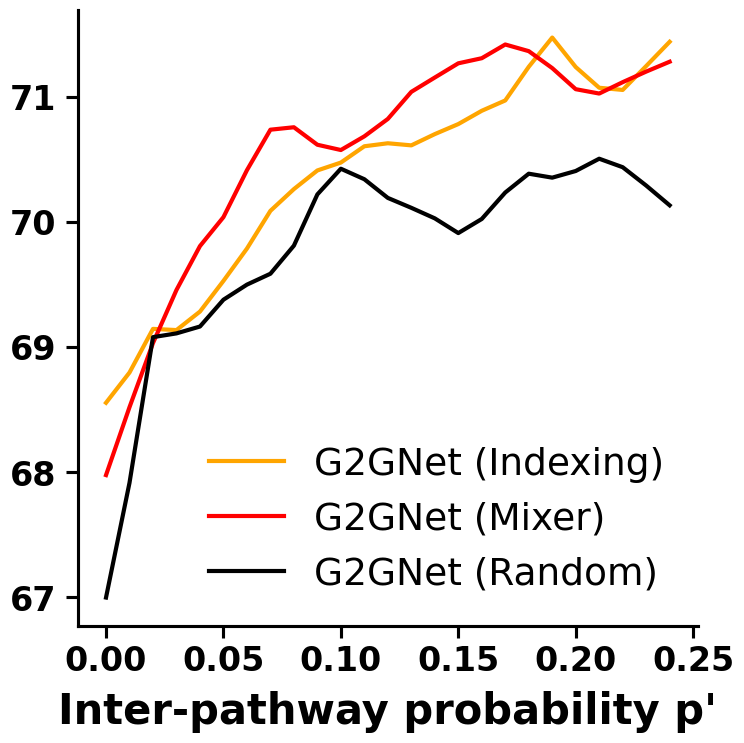}
        %Grouping techniques}
        \label{fig:mix_res}
    \end{subfigure}
    \hfill
    \begin{subfigure}[b]{0.19\textwidth}
        \centering
        \caption{}
        \includegraphics[width=\linewidth]{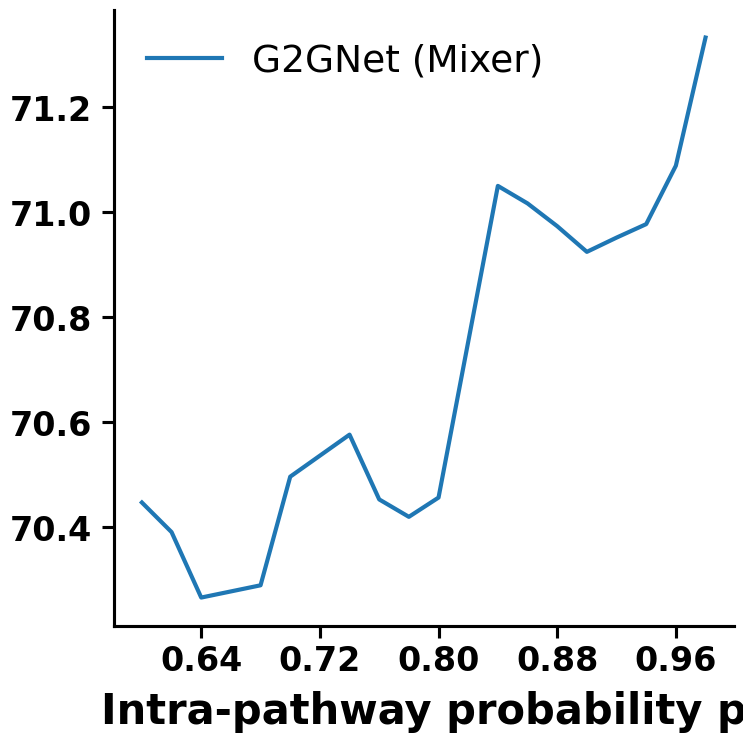}
        %\caption{Impact of probability \(p\).}
        \label{fig:p}
    \end{subfigure}
    \hfill
    \begin{subfigure}[b]{0.19\textwidth}
        \centering
        \caption{}
        \includegraphics[width=\linewidth]{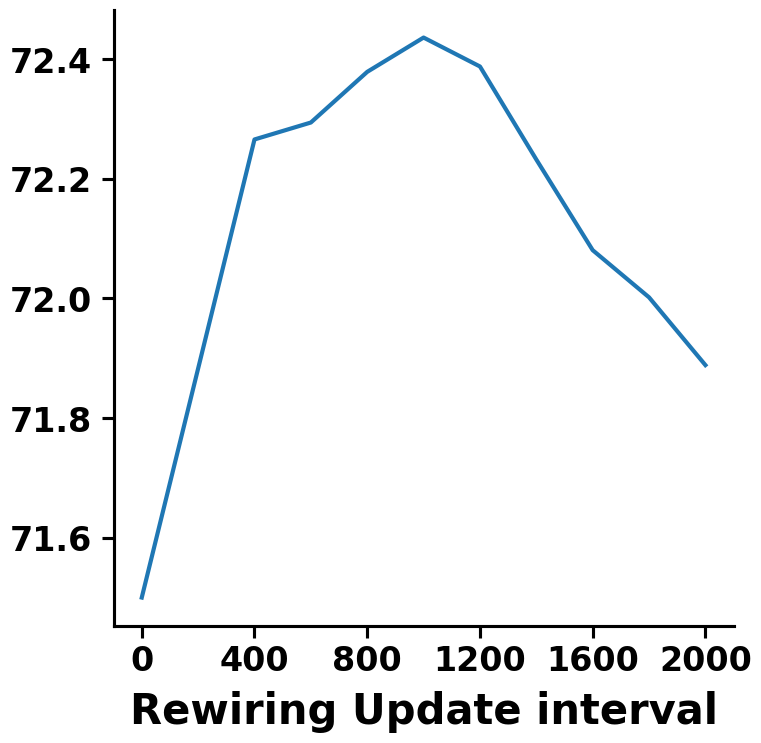}
        %\caption{Effect of the update interval.}
        \label{fig:dt}
    \end{subfigure}
    \hfill
    \begin{subfigure}[b]{0.19\textwidth}
        \centering
        \caption{}
        \includegraphics[width=\linewidth]{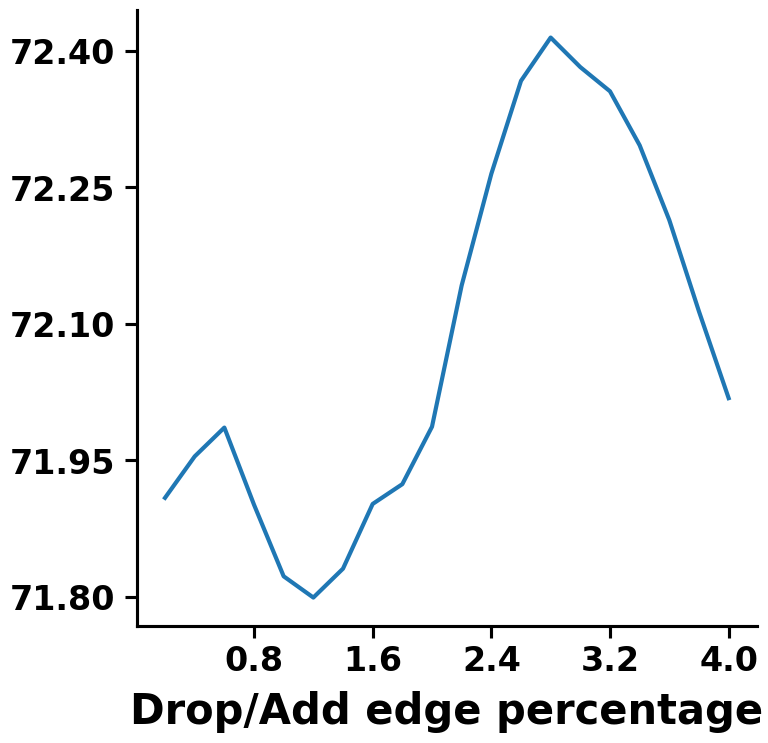}
        %\caption{Effect of drop/add percentage.}
        \label{fig:per}
    \end{subfigure}

    \caption{\textbf{Experimental results on CIFAR-10 }(a) As the inter-pathway probability \(p'\) of connecting sparsely connected groups, and thus the number of edges, increases, accuracy improves and plateaus around 70.8\% for \(p' \in [0.08, 0.15]\), corresponding to approximately 18\% to 25\% sparsity. Compared to Erdős-Rényi (ER) graphs of equal number of edges, our structure bias consistently yields higher accuracy. (b) The Mixer technique outperforms grouping by indexing across a wide range of \(p'\) values, while random grouping significantly underperforms. (c) Increasing the intra-pathway probability \(p\) of connecting densely connected groups leads to higher accuracy; we therefore set \(p=1\) in all experiments.
    (d) Effect of the rewiring update interval: accuracy is evaluated as a function of the number of training iterations between edge removal and addition steps. The best performance is observed when the update occurs every 1000 iterations. (e) Effect of the \(\%\) of drop/add edges: accuracy is measured for different proportions of weights pruned and regrown, with the update interval fixed at 1000 iterations. The optimal performance is achieved around a 2.5\% update rate.
    Reported values are mean accuracies over five runs with identical hyperparameters. For grouping strategy comparisons, the connectivity matrix initialization is kept fixed. }
    
    \label{fig:combined}
\end{figure*}

\textbf{Is the G2GNet structure bias effective?}  
To assess the benefit of our proposed structure, we compare it to several alternative architectures:  
\begin{itemize}
    \item \textit{Fully-connected baseline:} We replace our G2GNet sparse network with fully-connected ones, either keeping the same width for each layer or adjusting the width to match the total number of parameters. Our proposed structure achieves higher accuracy when having equal number of parameters and similar or better accuracy when having \(~25\%\) of the parameters across our benchmarks.
    \item \textit{Random sparsity (Erdős–Rényi):} We construct uniformly random sparse layers, dropping edges with probability \(\tilde{p}\) such that the total number of parameters matches ours. Again, our modular structure, which favors local connections, outperforms this random baseline, see Fig. \ref{fig:er_pprime}.
\end{itemize}

These results are summarized in Table~1.

\textbf{Understanding hyperparameters}  
We now examine the impact of the key hyperparameters of our architecture on accuracy, using the CIFAR-10 dataset. Figure \ref{fig:p} shows that increasing the intra-pathway probability \(p\) consistently improves accuracy. For this reason, we adopt \(p = 1\) as the default setting. Figure \ref{fig:er_pprime} illustrates the effect of the sparse inter-pathway probability \(p'\). Interestingly, accuracy exhibits a convex relationship with \(p'\): at very low \(p'\), where ensembles are largely disconnected, accuracy drops. Conversely, at higher \(p'\), where the network comes closer to the fully-connected one, accuracy saturates and slightly decreases due to the loss of the structural inductive bias. The best performance - sparsity trade-off is achieved for \(p' \in [0.08, 0.15]\), where the network remains sparse but inter-ensemble communication is preserved. 

\textbf{Comparing grouping techniques}  
In Figure \ref{fig:mix_res}, we also report the impact of the different grouping techniques described in Section~\ref{sec:gr}. We found that the Mixer technique performs best, as it enables dense modules to process information from diverse regions of the input image. Interestingly, when grouping is random, the information flow lacks any structure related to the spatial proximity of the input, and the results deteriorate to levels like the Erdős–Rényi connectivity pattern. 
%This further highlights the importance of incorporating an inductive bias in the way connections are structured.
\begin{comment}
\begin{figure}
    \centering
    \includegraphics[width=0.5\linewidth]{mixer_group.png}
    \caption{Comparing different grouping techniques: Accuracy on CIFAR-10 for different values of \(p'\). }
    \label{fig:mix_res}
\end{figure}
\end{comment}

\subsection{Sparse Training Evaluation}

Table~\ref{tab:dstraining-merged} shows the performance of various dynamic sparse training (DST) strategies applied to G2GNet on the CIFAR-10 and CIFAR-100 datasets. We evaluated all combinations of pruning (edge removal) criteria and the two growth (edge addition) methods described in Section~\ref{sec:meth}. Specifically, edges are pruned using one of the following strategies: (i) random selection, (ii) magnitude-based pruning (following the SET criterion~\cite{Mocanu}), or (iii) our proposed Hebbian-based criterion. Edges are then regrown either \textit{randomly }or using the \textit{Hebbian criterion}, which prioritizes connections between neurons with highly correlated activation patterns.

We did not consider DST methods that rely on dense gradient storage~\cite{riggl, topk}, as they require maintaining gradient information for inactive connections, which undermines the computational benefits of sparsity.

Empirically, we found that DST strategies consistently improve the performance of G2GNet on CIFAR-10, and on CIFAR-100 when the Hebbian growth criterion is employed. Among the various combinations, the Hebbian criterion generally performs best for edge addition, consistently outperforming random growth. The most effective configuration involves pruning by magnitude (SET) and growing with the Hebbian criterion on CIFAR-10, while on CIFAR-100, the best results are achieved by using the Hebbian criterion for both pruning and growth.

We then fixed the DST configuration to use Hebbian-based criteria for both pruning and growth, and conducted a sensitivity analysis on two key hyperparameters: the \textit{weight update interval} and the \textit{proportion of connections modified per update} (see Figures~\ref{fig:dt}, \ref{fig:per}). In line with prior work~\cite{fantastic}, we found that performance is optimized when updates occur every 1000 training iterations, with an update rate of approximately 2.5\%. When the update interval is too short or the update rate too high, the network struggles to adapt and train the newly-added parameters effectively. On the other hand, if updates are too infrequent or the update rate too low, the impact of the rewiring mechanism diminishes.

\begin{comment}
    
\begin{table}[ht]
\centering
\caption{Test accuracies (\%) on CIFAR-10 for combinations of edge \textbf{removal} (rows) and edge \textbf{addition} (columns) in Dynamic Sparse Training (DST). Averaged over five runs with fixed initial topology.}
\label{tab:dstraining-cifar10}
\begin{tabular}{llcc}
\toprule
&  & \multicolumn{2}{c}{\textbf{Addition Strategy}} \\
& \textbf{Removal Strategy} & \textbf{Random} & \textbf{Hebbian (ours)} \\
\midrule
& SET (Magnitude-based) \cite{Mocanu} & 71.6 & \textbf{72.7} \\
& Random                              & \textbf{72.5} & 72.0 \\
& Hebbian (ours)                      & 71.5 & 72.5 \\
\midrule
\multicolumn{3}{l}{\textbf{Static baseline (no updates)}}  71.4 \\
\bottomrule
\end{tabular}
\end{table}

\begin{table}[ht]
\centering
\caption{Test accuracies (\%) on CIFAR-100 for combinations of edge \textbf{removal} (rows) and edge \textbf{addition} (columns) in Dynamic Sparse Training (DST). Averaged over five runs with fixed initial topology.}
\label{tab:dstraining-cifar100}
\begin{tabular}{llcc}
\toprule
&  & \multicolumn{2}{c}{\textbf{Addition Strategy}} \\
& \textbf{Removal Strategy} & \textbf{Random} & \textbf{Hebbian (ours)} \\
\midrule
& SET (Magnitude-based) \cite{Mocanu} & 41.8 & 42.3 \\
& Random                              & 41.8 & 42.28 \\
& Hebbian (ours)                      & 41.4 & \textbf{42.7} \\
\midrule
\multicolumn{3}{l}{\textbf{Static baseline (no updates)}}  42.2 \\
\bottomrule
\end{tabular}
\end{table}
\end{comment}
\begin{table}[ht]
\centering
\caption{Test accuracies (\%) on CIFAR-10 and CIFAR-100 for combinations of edge \textbf{removal} (rows) and edge \textbf{addition} (columns) in Dynamic Sparse Training (DST). Averages over five runs with fixed initial topology are reported.}
\label{tab:dstraining-merged}
\adjustbox{max width=\linewidth}{
\begin{tabular}{llcccc}
\toprule
\multicolumn{2}{c}{} & \multicolumn{2}{c}{\textbf{CIFAR-10}} & \multicolumn{2}{c}{\textbf{CIFAR-100}} \\
\multicolumn{2}{c}{} & \multicolumn{4}{c}{\textbf{Addition Strategy}} \\
& \textbf{Removal Strategy} & \textbf{Random} & \textbf{Hebbian} & Random & \textbf{Hebbian} \\
\midrule
& SET (Magnitude-based) \cite{Mocanu} & 71.6 & \textbf{72.7} & 41.8 & 42.3 \\
& Random                               & \textbf{72.5} & 72.0 & 41.8 & 42.28 \\
& Hebbian (ours)                       & 71.5 & 72.5 & 41.4 & \textbf{42.7} \\
\midrule
\multicolumn{2}{l}{\textbf{Static baseline (no updates)}} & \multicolumn{2}{c}{71.4} & \multicolumn{2}{c}{42.2} \\
\bottomrule
\end{tabular}
}
\end{table}

\begin{comment}
\begin{figure*}[ht]
    \centering
    \begin{subfigure}[b]{0.48\linewidth}
        \centering
        \includegraphics[height=5cm]%[width=\linewidth]
        {dt.png}
        \caption{Effect of the update interval}
        \label{fig:dt}
    \end{subfigure}
    \hfill
    \begin{subfigure}[b]{0.48\linewidth}
        \centering
        \includegraphics[height=5cm]%[width=\linewidth]
        {per.png}
        \caption{Effect of the drop/add percentage}
        \label{fig:per}
    \end{subfigure}
    \caption{Sensitivity analysis of dynamic sparse training (DST) hyperparameters. (a) Effect of the update interval: accuracy is evaluated as a function of the number of training iterations between edge removal and addition steps. The best performance is observed when the update occurs every 1000 iterations. (b) Effect of the drop/add percentage: accuracy is measured for different proportions of weights pruned and regrown, with the update interval fixed at 1000 iterations. The optimal performance is achieved around a 2.5\% update rate.}

    \label{fig:sensitivity}
\end{figure*}
    
\end{comment}

\section{Discussion and Future Work} \label{sec:concl} 
Inspired by ensemble-to-ensemble information transmission observed in biological circuits of the mouse visual cortex, we proposed a structural bias for the feedforward layers of artificial neural networks. We refer to this architecture as \textbf{G2GNet}: a sparse network that facilitates information flow through static, coupled group pairs across adjacent layers, while still allowing \textit{controlled information leakage} between uncoupled group pairs. This design introduces structured pathways through the depth of the network, which can be columnar, randomly shaped, or mixed, depending on the grouping strategy.

The initial connectivity pattern imposed by G2GNet may not contain the “winning ticket”, a subnetwork that performs as well as the full network, as described by the Lottery Ticket Hypothesis. To address this, we incorporate a dynamic edge rewiring mechanism that maintains a fixed number of pruned and newly-activated edges while adapting connectivity during training. Central to this mechanism is our proposed \textbf{Hebbian-inspired criterion}, which adds or prunes edges based on the activation correlation between connected neurons. Our approach yields strong performance on computer vision benchmarks and enables \textbf{structured sparsity}, a form of sparsity that is more compatible with current hardware acceleration.

We demonstrate that: \textbf{i)} G2GNet is an effective method for matrix sparsification; \textbf{ii)} allowing communication between distinct pathways improves performance; and \textbf{iii)} functional connectivity patterns, and more specifically, pairwise correlations of the activation of neurons can guide the edge rewiring in ANNs, enhancing the efficiency.
%and \textbf{iv)} insights from biological networks can inform the design of efficient artificial architectures.

While our method is \textit{only conceptually} inspired by biological principles, such as ensemble-to-ensemble communication and information flow, it opens promising directions for future research. In particular, the notion of functional groups in biological networks could be extended to dynamic group formation in ANNs, driven by input data, internal network state, or even as a means for network training and avoiding of catastrophic forgetting.
Currently, our experiments focus on feedforward layers, which limits the complexity of tasks evaluated. However, the underlying structural bias of G2GNet is applicable to convolutional layers as well as to the query, key, and value matrices of attention mechanisms in transformer-based architectures. Moreover, part of our on-going efforts includes the analysis of these ideas on SNNs and an in-depth examination of mixer-like grouping strategies.
G2GNet provides an example of how the organizational principles identified in neuroscience (e.g., in \cite{pap})  may inform the design of ANNs, providing also a simulation tool for testing of neuro-computational
hypotheses.
%%%%%%%%%%%%%%%
\section*{Acknowledgments}
This work has been partially supported by the project MIS 5154714 of the National Recovery and Resilience Plan Greece 2.0 funded by the European Union under the NextGenerationEU Program.
It has also received funding from the European Union’s Horizon 2020 research and innovation program under the Marie Skłodowska-Curie grant agreement No 101007926 neuronsXnets as well as from the Hellenic Foundation Research Institute (HFRI) with the neuron-AD project number 04058 and project number 2285 (PI: Maria Papadopouli). 
\begin{comment}
%The review process is double-bind, so at this phase, we should not have ACK
%\section*{Acknowledgment}
\section*{Acknowledgments}

This work has received funding from the European Union’s Horizon 2020 research and innovation program under the Marie Skłodowska-Curie grant agreement No 101007926 as well as from the Hellenic Foundation Research Institute (HFRI) with the neuron-AD project number 04058 and neuronXnet project number 2285 (PI: Maria Papadopouli). 
\end{comment}
\bibliographystyle{IEEEtran}
%\bibliography{references}

%\subfile{appendix.tex}

\end{document}